\documentclass{article}

     \PassOptionsToPackage{numbers, compress}{natbib}

     \usepackage[preprint]{neurips_2020}

\usepackage[utf8]{inputenc} 
\usepackage[T1]{fontenc}    
\usepackage{hyperref}       
\usepackage{url}            
\usepackage{booktabs}       
\usepackage{amsfonts}       
\usepackage{nicefrac}       
\usepackage{microtype}      
\usepackage{amsthm}
\usepackage{amssymb}
\usepackage{amsmath}
\usepackage{url}
\usepackage{graphicx}
\usepackage{amsfonts}
\usepackage{amsfonts}
\usepackage{amssymb}
\usepackage{pifont}

\newtheorem{prop}{Proposition}
\newtheorem{rem}{Remark}
\newtheorem{thm}{Theorem}

\newtheorem{exmp}{Example}
\newtheorem{cor}{Corollary}
\newtheorem{ass}{Assumption}
\def\norm#1{\Vert#1\Vert}
\DeclareMathOperator{\ph}{Phase}

\newcommand{\cmark}{\ding{51}}%
\newcommand{\xmark}{\ding{55}}%

\newcommand*\samethanks[1][\value{footnote}]{\footnotemark[#1]}

\title{Implicit bias of deep linear networks in the large learning rate phase}

\author{%
  Wei~Huang \thanks{Both authors contributed equally to this work.}\\
  University of Technology Sydney, Australia \\
  \texttt{wei.huang-6@student.uts.edu.au} \\
   \And
   Weitao Du   \samethanks[1] \\
   Northwestern University, USA \\
   \texttt{weitao.du@northwestern.edu} \\
   \And
   Richard Yi Da Xu \\
   University of Technology Sydney, Australia \\
   \texttt{YiDa.Xu@uts.edu.au} \\
   \And
   Chunrui Liu \\
   University of Technology Sydney, Australia \\
   \texttt{chunrui.liu@student.uts.edu.au}
}

\begin{document}

\maketitle

\begin{abstract}

Most theoretical studies explaining the regularization effect in deep learning have only focused on gradient descent with a sufficient small learning rate or even gradient flow (infinitesimal learning rate). Such researches, however, have neglected a reasonably large learning rate applied in most practical applications. In this work, we characterize the implicit bias effect of deep linear networks for binary classification using the logistic loss in the large learning rate regime, inspired by the seminal work by Lewkowycz et al. \cite{lewkowycz2020large} in a regression setting with squared loss. They found a learning rate regime with a large stepsize named the {\it catapult phase}, where the loss grows at the early stage of training and eventually converges to a minimum that is flatter than those found in the small learning rate regime. We claim that depending on the separation conditions of data, the gradient descent iterates will converge to a flatter minimum in the catapult phase. We rigorously prove this claim under the assumption of degenerate data by overcoming the difficulty of the non-constant Hessian of logistic loss and further characterize the behavior of loss and Hessian for non-separable data. Finally, we demonstrate that flatter minima in the space spanned by non-separable data along with the learning rate in the catapult phase can lead to better generalization empirically.

\end{abstract}

\section{Introduction}
\label{intro}

Deep neural networks have been achieving a variety of successes in both supervised and unsupervised learning. The theoretical understanding of the mechanism underlying deep learning power is continually refining and expanding by researchers. Remarkably, recent progress in deep learning theory has shown that over-parameterized networks are capable of achieving very low or even zero training error through gradient descent based optimization \cite{jacot2018neural,allen2019convergence,du2018gradient,chizat2018global,zou2018stochastic}. Meanwhile, these over-parameterized networks can generalize well to the test set, known as the double descent phenomenon \cite{nakkiran2019deep}. 

How can over-parameterized networks survive from the over-fitting problem given that model parameters are far more than the sample size of training data or the input dimension? Among existing explanations for this question, a promising answer is the implicit bias \cite{soudry2018implicit} or implicit regularization \cite{neyshabur2014search}. Specifically, a large family of works studying the exponential tailed losses, including logistic and exponential loss, has reported a strong regularization resulted from the maximum margin \cite{soudry2018implicit,gunasekar2018characterizing,ji2019implicit,nacson2019lexicographic,lyu2019gradient}. 

Notably, all theoretical results associated with implicit bias are assuming the learning rate is sufficiently small. The theoretical understanding of optimization and generalization properties remains limited when the learning rate goes beyond the small learning rate setting. Furthermore, adopting a learning rate annealing scheme with a large initial learning rate often achieves better performance in practice \cite{he2016deep,zagoruyko2016wide}. Lewkowycz et al. \cite{lewkowycz2020large} shed light on the large learning rate phase by observing a distinct phenomenon that the local curvature of the loss landscape drops significantly in the large learning rate phase and thus typically can obtain the best performance. 

By following \cite{lewkowycz2020large}, we characterize the gradient descent training in terms of three learning rate regimes or phases: (i) {\bf lazy phase} $\eta<\eta_0$, when the learning rate is small, the dynamics of neural networks its linearized dynamics regime, where model converges to a nearby point in parameter space which is called {\it lazy training} and characterized by the neural tangent kernel  \cite{jacot2018neural,arora2019exact,yang2019scaling,huang2019dynamics,allen2019convergence,du2018gradient}. (ii) {\bf catapult phase} $\eta_0<\eta<\eta_1$, the loss grows at the beginning and then drops until it converges to the solution with a flatter minimum. (iii) {\bf divergent phase} $ \eta>\eta_1$, the loss diverges, and the model does not train. The importance of the {\it catapult phase} rises because the {\it lazy phase} is in general detrimental to generalization and does not explain the practically observed power of deep learning \cite{allen2019can,chizat2019lazy}.

While the phenomenon of three learning rate phases is reported in a regression setting with mean squared error (MSE) loss, it remains unclear whether this can be extended to cross-entropy (logistic) loss. To fill this gap, we exam the effect of the large learning rate on deep linear networks with logistic and exponential loss. Contrary to MSE loss, the characterization of gradient descent with logistic loss concerning learning rate is associated with separation conditions of data. Finally, we summarize our contribution as follows: 
\begin{itemize}
\item According to separation conditions of data, we characterize the dynamics of gradient descent with logistic and experiential loss corresponding to the learning rate. We find the gradient descent iterates converge to a flatter minimum in the catapult phase when the data is non-separable. Such three learning rate phases do not apply to the linearly separable data since the optimum is off at infinity.
\item Our theoretical analysis ranges from linear predictor to one hidden layer network. By comparing the convex optimization characterized by Theorem \ref{thm:linear_predictor_degenerate} and non-convex optimization characterized by Theorem \ref{thm:linear_network_degenerate} in terms of the learning rate, we show that the {\it catapult phase} is a unique phenomenon for non-convex optimization. 
\item We find that in practical classification tasks, the best generalization results tend to occur in the {\it catapult phase}. Given the fact that the infinite-width analysis ({\it lazy training}) does not fully explain the empirical power of deep learning, our result can be used to fill this gap partially.
\end{itemize}

\section{Related Work}
\label{rela}

\paragraph{Implicit bias of gradient methods.} Since the seminal work from \cite{soudry2018implicit}, implicit bias has led to a fruitful line of research. Works along this line have treated linear predictors \cite{gunasekar2018characterizing,ji2019implicit,ali2020implicit}; deep linear networks with a single output \cite{nacson2018convergence,gunasekar2018implicit,ji2018gradient} and multiple outputs \cite{radhakrishnan2020balancedness}; homogeneous networks (including ReLU, max pooling activation) \cite{nacson2019lexicographic,lyu2019gradient,ji2020directional}; ultra wide networks \cite{chizat2020implicit,oymak2018overparameterized}; matrix factorization \cite{razin2020implicit}. Notably, these studies adopt gradient flow (infinitesimal learning rate) or a sufficient small learning rate. 

\paragraph{Neural tangent kernel.} Recently, we have witnessed exciting theoretical developments in understanding the optimization of ultra wide networks, known as the neural tangent kernel (NTK) \cite{jacot2018neural,arora2019exact,yang2019scaling,huang2019dynamics,allen2019convergence,du2018gradient,zou2018stochastic}. It is shown that in the infinite-width limit, NTK converges to an explicit limiting kernel, and it stays constant during training. Further, Lee et al. \cite{lee2019wide} show that gradient descent dynamics of the original neural network fall into its linearized dynamics regime in the NTK regime. In addition, the NTK theory has been extended to various architecture such as orthogonal initialization \cite{huang2020neural}, convolutions \cite{arora2019exact,li2019enhanced}, graph neural networks \cite{du2019graph}, attention \cite{hron2020infinite}, and batch normalization \cite{jacot2019freeze} (see \cite{yang2020tensor} for a summary). The constant property of NTK during training can be regarded as a special case of the implicit bias, and importantly, it is only valid in the small learning rate regime. 

\paragraph{Large learning rate and logistic loss} A large learning rate with SGD training is often set initially to achieve good performance in deep learning empirically  \cite{krizhevsky2012imagenet,he2016deep,zagoruyko2016wide}. Existing theoretical explanation of the benefit of the large learning rate contributes to two classes. One is that a large learning rate with SGD leads to flat minima \cite{keskar2016large,jiang2019fantastic,lewkowycz2020large}, and the other is that the large learning rate acts as a regularizer \cite{li2019towards}. Especially, Lewkowycz et al. \cite{lewkowycz2020large} find a large learning rate phase can result in flatter minima without the help of SGD for mean squared loss. In this work, we ask whether the large learning rate still has this advantage with logistic loss. We expect a different outcome because the logistic loss is sensitive to the separation conditions of the data, and the loss surface is different from that of MSE loss \cite{nitanda2019gradient}.

\section{Background}

\subsection{Setup}

Consider a dataset $\{ x_i, y_i \}_{i=1}^n$, with inputs $x_i \in \mathbb{R}^d $ and binary labels $y_i \in \{-1,1 \}$. The empirical risk of classification task follows the form,
  \begin{equation} \label{eq:propagation}
    \mathcal{L}  =  \frac{1}{n}\sum_{i=1}^n \ell (f(x_i) y_i),
  \end{equation}
where $f(x_i)$ is the output of model corresponds to the input $x_i$, and $\ell(\cdot)$ is loss function. In this work, we study two exponential tail losses which are exponential loss $\ell_{\exp}(u) = \exp(-u)$ and logistic loss $\ell_{\log}(u) = \log(1+\exp(-u))$. The reason we look at these two losses together is that they are jointly considered in the realm of implicit bias by default \cite{soudry2018implicit}. We adopt gradient descent (GD) updates with learning rate $\eta$ to minimize empirical risk,
  \begin{equation} \label{eq:propagation}
    w_{t+1} = w_t-\eta \nabla \mathcal{L}(w_t) = w_t-\eta \sum_{i=1}^n \ell'(f(x_i) y_i),
  \end{equation}
where $w_t$ is the parameter of the model at time step $t$.

\subsection{Separation Conditions of Dataset}

It is known that landscapes of cross-entropy loss on linearly separable data and non-separable data are different. Thus the separation condition plays a crucial role in understanding the dynamics of gradient descent in terms of learning rate. To build towards this, we define the two classes of separation conditions and review existing results for loss landscapes of a linear predictor in terms of separability. 

\begin{ass}
The dataset is linearly separable, i.e. there exists a separator $w_\ast$ such that $\forall i: w^T_\ast x_i y_i > 0$.
\label{ass:1}
\end{ass}

\begin{ass}
The dataset is non-separable, i.e. there is no separator $w_\ast$ such that $\forall i: w^T_\ast x_i y_i > 0$.
\label{ass:2}
\end{ass}

\paragraph{Linearly separable.} Consider the data under assumption \ref{ass:1}, one can examine that the loss of a linear predictor, i.e., $f(x) = w^T x$, is $\beta$-smooth convex with respect to $w$, and the global minimum is at infinity. The implicit bias of gradient descent with a sufficient small learning rate ($\eta < \frac{2}{\beta}$) in this phase has been studied by \cite{soudry2018implicit}. They show the predictor converges to the direction of the maximum margin (hard margin SVM) solution, which implies the gradient descent method itself will find a proper solution with an implicit regularization instead of picking up a random solver. If one increases the learning rate until it exceeds $\eta<\frac{2}{\beta}$, then the result of converging to maximum margin will not be guaranteed, though loss can still converge to a global minimum. 

\paragraph{Non-separable.} Suppose we consider the data under assumption \ref{ass:2}, which is not linearly separable. The empirical risk of a linear predictor on this data is $\alpha$-strongly convex, and the global minimum is finite. In this case, given an appropriate small learning rate ($\eta < \frac{2}{\beta}$), the gradient descent converges towards the unique finite solution. When the learning rate is large enough, i.e., $\eta>\frac{2}{\alpha}$, we can rigorously show that gradient descent update with this large learning rate leads to risk exploding or saturating. 

We formally construct the relationship between loss surfaces and learning dynamics of gradient descent with respect to different learning rates on the two classes of data throng the following proposition,
\begin{prop}\label{prop:linear_predictor}
For a linear predictor $f = w^T x $, along with a loss $\ell \in \{ \ell_{\exp}, \ell_{\log} \}$. 
\begin{enumerate}
\item [1] 
Under Assumption \ref{ass:1}, the empirical loss is $\beta$-smooth. Then the gradient descent with constant learning rate $\eta < \frac{2}{\beta}$ never increase the risk, and empirical loss will converge to zero:
\[
\mathcal{L}(w_{t+1}) - \mathcal{L}(w_t) \le 0, ~~~ \lim_{t \rightarrow \infty} \mathcal{L}(w_t) = 0, ~~~with~~~  \eta < \frac{2}{\beta}
\]
\item [2] 
Under Assumption \ref{ass:2}, the empirical loss is $\beta$-smooth and $\alpha$-strongly convex, where $\alpha \le \beta$. Then the gradient descent with a constant learning rate $\eta < \frac{2}{\beta}$ never increases the risk, and empirical loss will converge to a global minimum. On the other hand, the gradient descent with a constant learning rate $\eta > \frac{2}{\alpha}$ never decrease the risk, and empirical loss will explode or saturate:
\[
\begin{aligned}
&\mathcal{L}(w_{t+1}) - \mathcal{L}(w_{t}) \le 0, ~~~ \lim_{t \rightarrow \infty} \mathcal{L}(w_t) = G_0,~~~ with~~~  \eta < \frac{2}{\beta} \\
&\mathcal{L}(w_{t+1}) - \mathcal{L}(w_t) \ge 0, ~~~ \lim_{t \rightarrow \infty} \mathcal{L}(w_t) = G_1,~~~ with~~~  \eta > \frac{2}{\alpha}
\end{aligned}
\]
where $G_0$ is the value of a global minimum while $G_1 = \infty$ for exploding situation or $G_0 <G_1 < \infty$ when saturating.
\end{enumerate}
\end{prop}


\section{Theoretical results}
\subsection {Convex Optimization}

It is known that the Hessian of the logistic and exponential loss with respect to the linear predictor is non-constant. Moreover, the estimated $\beta$-smooth convexity and $\alpha$-strongly convexity vary across different finite bounded subspace. As a result, the learning rate threshold in Proposition \ref{prop:linear_predictor} is not detailed in terms of optimization trajectory. However, we can obtain more elaborate thresholds of the learning rate for linear predictor by considering the {\it degeneracy assumption}:
\begin{ass}
 The dataset contains two data points where they have same feature and opposite label, that is
$$(x_1 = 1, y_1 = 1)\ \  \ \text{and}\ \ \ (x_2 = 1,y_2 = -1).$$
\label{ass:3}
\end{ass}

We call this assumption the {\it degeneracy assumption} since the features from opposite label degenerate. Without loss of generality, we simplify the dimension of data and fix the position of the feature. Note that this assumption can be seen as a special case of non-separable data. There is a work theoretically characterizing general non-separable data \cite{ji2019implicit}, and we leave the analysis of this setting for the large learning rate to future work. Thanks to the symmetry of the risk function in space at the basis of {\it degeneracy assumption}, we can construct the exact dynamics of empirical risk with respect to the whole learning rate space.
\begin{thm}\label{thm:linear_predictor_degenerate}
For a linear predictor $f = w^T x $ equipped with exponential (logistic) loss under assumption \ref{ass:3}, there is a critical learning rate that separates the whole learning rate space into two (three) regions. The critical learning rate satisfies $$\mathcal{L}'(w_0) = -\mathcal{L}'(w_0-\eta_{\rm critical} \mathcal{L}'(w_0)),$$ where $w_0$ is the initial weight. Moreover,  
\begin{enumerate}
\item [1] 
For exponential loss, the gradient descent with a constant learning rate $\eta < \eta_{\rm critical}$ never increases loss, and the empirical loss will converge to the global minimum. On the other hand, the gradient descent with learning rate $ \eta = \eta_{\rm critical}$ will oscillate. Finally, when the learning rate $  \eta > \eta_{\rm critical}$, the training process never decreases the loss and the empirical loss will explode to infinity:
\[
\begin{aligned}
& \mathcal{L}(w_{t+1}) - \mathcal{L}(w_t) < 0,  ~~~ \lim_{t \rightarrow \infty} \mathcal{L}(w_t) = 1,~~~ with~~~  \eta < \eta_{\rm critical}, \\
& \mathcal{L}(w_{t+1}) - \mathcal{L}(w_t) = 0,  ~~~ \lim_{t \rightarrow \infty} \mathcal{L}(w_t) = \mathcal{L}(w_0),~~~ with~~~  \eta = \eta_{\rm critical},  \\
& \mathcal{L}(w_{t+1}) - \mathcal{L}(w_t) > 0,  ~~~ \lim_{t \rightarrow \infty} \mathcal{L}(w_t) = \infty,~~~ with~~~  \eta > \eta_{\rm critical}.  \\
\end{aligned}
\]
\item [2] 
For logistic loss, the critical learning rate satisfies a condition: $\eta_{\rm critical} > 8$. The gradient descent with a constant learning rate $\eta < 8$ never increases the loss, and the loss will converge to the global minimum. On the other hand, the loss along with a learning rate $8 \le \eta <\eta_{\rm critical} $ will not converge to the global minimum but oscillate. Finally, when the learning rate $  \eta > \eta_{\rm critical}$, gradient descent never decreases the loss, and the loss will saturate:
\[
\begin{aligned}
& \mathcal{L}(w_{t+1}) - \mathcal{L}(w_t) < 0,   ~~\lim_{t \rightarrow \infty} \mathcal{L}(w_t) = \log(2),~with~ \eta < 8, \\
& \mathcal{L}(w_{t+1}) - \mathcal{L}(w_t) \le 0,   ~~\lim_{t \rightarrow \infty} \mathcal{L}(w_t) = \mathcal{L}(w_\ast) < \mathcal{L}(w_0), ~with~  8 \le \eta < \eta_{\rm critical},  \\
& \mathcal{L}(w_{t+1}) - \mathcal{L}(w_t) \ge 0,   ~~\lim_{t \rightarrow \infty} \mathcal{L}(w_t) = \mathcal{L}(w_\ast) \ge \mathcal{L}(w_0),~with~    \eta \ge \eta_{\rm critical}.  \\
\end{aligned}
\]
where $w_\ast$ satisfies $-w_\ast = w_\ast - \frac{\eta}{2} \frac{\sinh(w_\ast)}{1+\cosh(w_\ast)}$.
\end{enumerate}
\end{thm}

\begin{rem}
The differences between the two loss is due to monotonicity of the loss. For exponential loss the function $|\mathcal{L}'(w_t)/w_t|$ is monotonically increasing with respect to $|w_t|$, while it is monotonically decreasing for logistic loss. 
\end{rem}

\begin{figure*}[t!]
\centering
  \centering
  \includegraphics[width=1.0\textwidth]{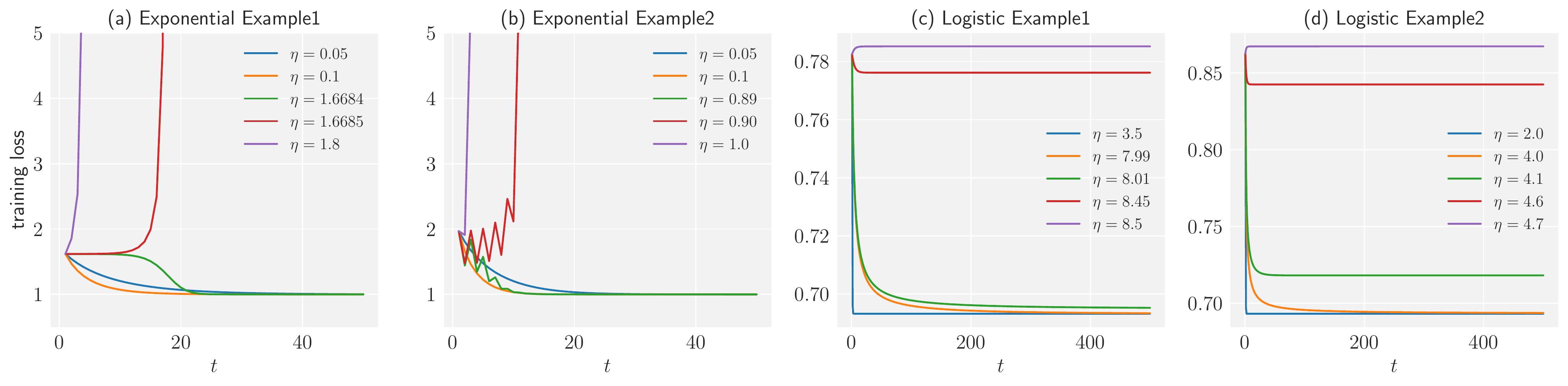} 
 \caption{Dependence of dynamics of training loss on the learning rate for linear predictor, with (a,b) exponential loss and (c,d) logistic loss on Example \ref{example:1} and \ref{example:2}. (a,c) The experimental learning curves are consistent with the theoretical prediction, and the critical learning rates are $\eta_{\rm critical}= 1.66843$ and $\eta_{\rm critical}= 8.485$ respectively.  (b,d) For non-separable data, the dynamics of training loss regarding learning rate for non-separable data are similar to those of degenerate case. Hence the critical learning rates can be approximated by $\eta_{\rm critical}= 0.895$ and  $\eta_{\rm critical}= 4.65$ respectively.}
 \label{fig:1}
\end{figure*}

We demonstrate the gradient descent dynamics with degenerate and non-separable case through the following example.
\begin{exmp}\label{example:1}
Consider optimizing $\mathcal{L}(w)$ with dataset  $\{ (x_1 = 1, y_1 = 1)\ \   \text{and}\ \ (x_2 = 1,y_2 = -1).\}$ using gradient descent with constant learning rates. Figure \ref{fig:1}(a,c) show the dependence of different dynamics on the learning rate $\eta$ for exponential and logistic loss respectively.
\end{exmp}

\begin{exmp}\label{example:2}
Consider optimizing $\mathcal{L}(w)$ with dataset  $\{ (x_1 = 1, y_1 = 1),\ \ (x_2 = 2,y_2 = -1)\ \ \text{and}\ \ (x_3 = -1,y_3 = 1).\}$ using gradient descent with constant learning rates. Figure \ref{fig:1}(b,d) show the dependence of different dynamics on the learning rate $\eta$ for exponential and logistic loss respectively. . 
\end{exmp}
\begin{rem}
The dataset considered here is an example of non-separable case, and the dynamics of loss behave similarly to those with Example \ref{example:1}. We use this example to show that our theoretical results on the degenerate data can be extended to the non-separable data empirically.
\end{rem}

\subsection{Non-convex Optimization}
\label{Two layer}

To investigate the relationship between the dynamics of gradient descent and the learning rate for deep linear networks, we consider linear networks with one hidden layer, and the information propagation in these networks is governed by,
  \begin{equation} \label{eq:propagation}
    f(x)  =  m^{-1/2} w^{(2)} w^{(1)} x,
  \end{equation}
where $m$ is the width, i.e. number of neurons in the hidden layer, $w^{(1)} \in \mathbb{R}^{m \times d}$ and $ w^{(2)} \in \mathbb{R}^{m}$ are the parameters of model. Taking the exponential loss as an example, the gradient descent equations at training step $t$ are,
  \begin{equation} \label{eq:gradient}
   \begin{aligned}
    w^{(1)}_{t+1}  &= w^{(1)}_{t} -\frac{1}{n} \frac{\eta} {m^{1/2}} (-e^{-y_\alpha f_t(x_\alpha)})  w^{(2)}_{t} x_{\alpha}y_\alpha, \\
    w^{(2)}_{t+1}  &= w^{(2)}_{t} -\frac{1}{n} \frac{\eta} {m^{1/2}} (-e^{-y_\alpha f_t(x_\alpha)})  w^{(1)}_{t} x_{\alpha}y_\alpha,
   \end{aligned}  
  \end{equation}
where we use the Einstein summation convention to simplify the expression and will apply this convention in the following derivation. 

We introduce the neural tangent kernel, an essential element for the evolution of output function in equation \ref{eq:evolution}. The neural tangent kernel (NTK) is originated from \cite{jacot2018neural} and formulated as,
  \begin{equation} \label{eq:ntk}
    \Theta_{\alpha \beta} = \frac{1}{m}  \sum_{p=1}^P \frac{\partial f(x_\alpha)}{\partial_{\theta_p}}\frac{\partial f(x_\beta)}{\partial_{\theta_p}}.
  \end{equation}
where $P$ is the number of parameters. For a two-layer linear neural network, the NTK can be written as,
  \begin{equation} \label{eq:ntk}
    \Theta_{\alpha \beta} = \frac{1}{mn} \big( (w^{(1)}x_{ \alpha}) ( w^{(1)}x_{ \beta})  + (w^{(2)})^2 (x_{\alpha} x_{\beta} ) \big).
  \end{equation}
Here we use normalized NTK which is divided by the number of samples $n$. Under the degeneracy assumption \ref{ass:3}, the loss function becomes  $\mathcal{L} = \cosh(m^{-1/2} w^{(2)} w^{(1)})$. Then the equation \ref{eq:gradient} reduces to 
\begin{equation}
\begin{aligned}
w^{(1)}_{t+1} = w^{(1)}_t -  \frac{\eta}{m^{1/2}} w^{(2)}_t  \sinh(m^{-1/2} w^{(2)}_t w^{(1)}_t), \\
w^{(2)}_{t+1} = w^{(2)}_t -  \frac{\eta}{m^{1/2}} w^{(1)}_t  \sinh(m^{-1/2} w^{(2)}_t w^{(1)}_t). \\
\end{aligned}
\end{equation}
The updates of output function $f_t$ and the eigenvalue of NTK $\lambda_t$, which are both scalars in our setting:
\begin{equation} \label{eq:evolution}
\begin{aligned}
f_{t+1}  & = f_t -  \eta \lambda_t \tilde {f_t}_{\rm exp} + \frac{\eta^2}{m} f_t \tilde {f^2_t}_{\rm exp}, \\
\lambda_{t+1} & = \lambda_t -  \frac{4\eta}{m} f_t \tilde {f_t}_{\rm exp} + \frac{\eta^2}{m} \lambda_t \tilde {f^2_t}_{\rm exp}. \\
\end{aligned}
\end{equation}
where $\tilde {f_t}_{\rm exp} :=  \sinh(f_t)$ while $\tilde {f_t}_{\rm log} :=  \frac{ \sinh(f_t)}{1+\cosh(f_t)}$ for logistic loss.

\begin{figure*}[t!]
\centering
  \centering
  \includegraphics[width=1.0\textwidth]{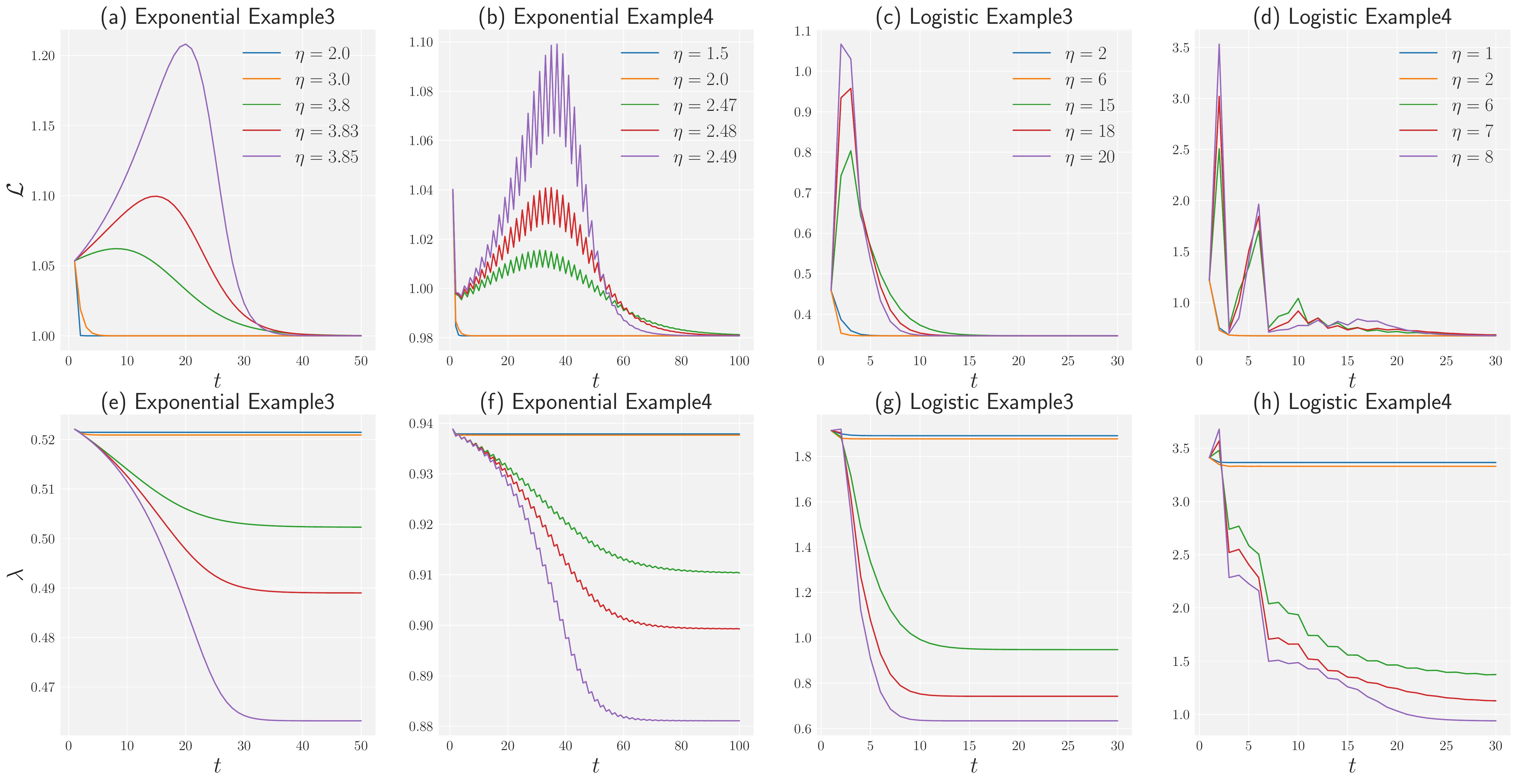}
 \caption{ Dependence of dynamics of training loss and maximum eigenvalue of NTK on the learning rate for one hidden layer linear network, with (a,b,e,f) exponential loss and (c,d,g,h) logistic loss on Example \ref{example:3} and \ref{example:4}. (a,b,c,d) In the large learning rate regime (the catapult phase), the loss increases at beginning and converges to a global minimum. (e,f,g,h) The maximum eigenvalue of NTK decreases rapidly to a fixed value which is lower than its initial position in the large learning regime (the catapult phase).} 
 \label{fig:2}
\end{figure*}

We have introduced the {\it catapult phase} where the loss grows at the beginning and then drops until converges to a global minimum. In the following theorem, we prove the existing of the {\it catapult phase} on the degenerate data with exponential and logistic loss.

\begin{thm}\label{thm:linear_network_degenerate}
Under appropriate initialization and assumption \ref{ass:3}, there exists a {\it catapult phase} for both the exponential and logistic loss. More precisely, when $\eta$ belongs to this phase, there exists a $T>0$ such that the output function $f_t$ and the eigenvalue of NTK $\lambda_t$ update in the following way:
\begin{enumerate}
	\item $\mathcal{L}_t$ keeps increasing when $t < T$.
	\item 
	After the $T$ step and its successors, the loss decreases, which is equivalent to:
	$$|f_{T+1}|>|f_{T+2}| \ge |f_{T+3}| \ge \dots.$$
	\item The eigenvalue of NTK keeps dropping after the $T$ steps:
	$$\lambda_{T+1} > \lambda_{T+2} \ge \lambda_{T+3} \ge \dots .$$	
\end{enumerate}
Moreover, we have the inverse relation between the learning rate and the final eigenvalue of NTK:
$	\lambda_\infty \le \lim_{t \rightarrow \infty} \frac{4f_t}{\eta \tilde {f_t}_{\rm exp}} $ with exponential loss, or $	\lambda_\infty \le \lim_{t \rightarrow \infty} \frac{4f_t}{\eta \tilde {f_t}_{\rm log} }$ with logistic loss.
\end{thm}


We demonstrate that {\it catapult phase} can be found in both degenerate and non-separable data through the following examples. The weights matrix is initialized by i.i.d. Gaussian distribution, i.e. $w^{(1)}, w^{(2)} \sim \mathcal{N}(0,\sigma^2_w)$. For exponential loss, we adopt the setting of $\sigma^2_w = 0.5$ and $m=1000$ while we set $\sigma^2_w = 1.0$ and $m=100$ for logistic loss. 

\begin{exmp}\label{example:3}
Consider optimizing $\mathcal{L}(w)$ using one hidden layer linear networks with dataset  $\{ (x_1 = [1,0], y_1 = 1)   \ \ \text{and}\ \ (x_2 = [1,0],y_2 = -1).\}$ and exponential (logistic) loss using gradient descent with constant learning rate. Figure \ref{fig:2}(a)(c)(e)(g) show how different choices of learning rate $\eta$ change the dynamics of loss function with exponential and logistic loss. 
\end{exmp}

\begin{exmp}\label{example:4}
Consider optimizing $\mathcal{L}(w)$ using one hidden layer linear networks with dataset  $\{ (x_1 = [1,1], y_1 = -1),\ \ (x_2 = [1,-1], y_1 = 1),\ \ (x_3 = [-1,-2], y_1 = 1)\ \ \text{and} \ \ (x_4 = [-1,1],y_4 = 1).\}$ and exponential (logistic) loss using gradient descent with constant learning rate. Figure \ref{fig:2}(b)(d)(f)(h) show how different choices of learning rate $\eta$ change the dynamics of loss function with exponential and logistic loss. 
\end{exmp}

As Figure \ref{fig:2} shows, in the {\it catapult phase}, the eigenvalue of NTK decreases to a lower value than its initial point, while it keeps unchanged in the {\it lazy phase} where the learning rate is small. For MSE loss, the lower value of neural tangent kernel indicates the flatter curvature given the training loss is low \cite{lewkowycz2020large}. Yet it is unknown whether aforementioned conclusion can be applied to exponential and logistic loss. Through the following corollary, we show that the Hessian is equivalent to the NTK when the loss converges to a global minimum for degenerate data.
\begin{cor} [Informal] \label{cor:hessian_ntk}
Consider optimizing $\mathcal{L}(w)$ with one hidden layer linear network under assumption \ref{ass:3} and exponential (logistic) loss using gradient descent with a constant learning rate. For any learning rate that loss can converge to the global minimal, the larger the learning rate, the flatter curvature the gradient descent will achieve at the end of training.
\end{cor}

\begin{figure*}[t!]
\centering
  \centering
  \includegraphics[width=1.0\textwidth]{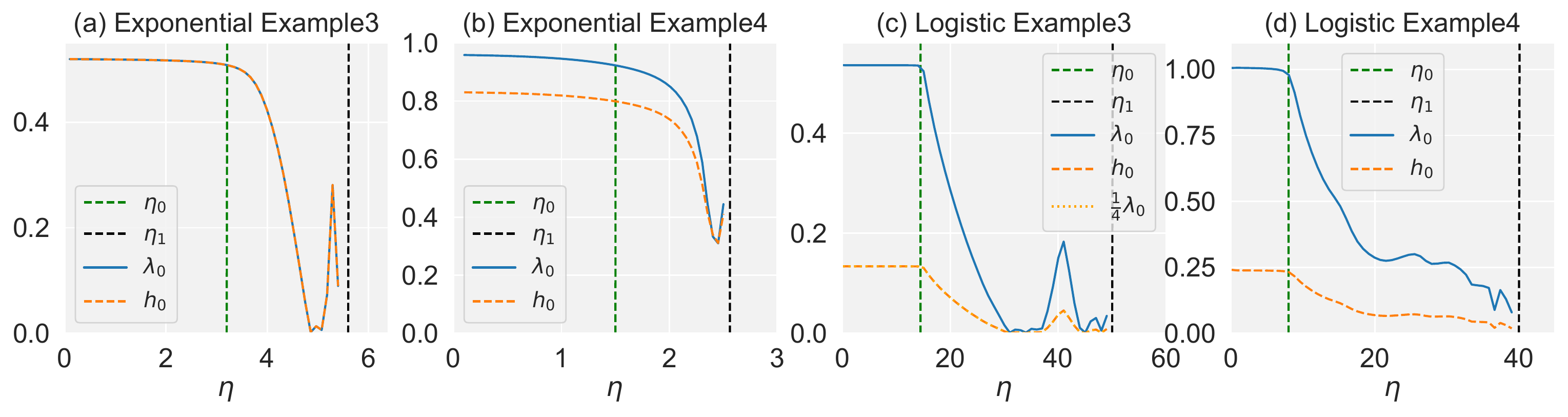}
 \caption{ Top eigenvalue of NTK ($\lambda_0$) and Hessian ($h_0$) measured at $t=100$ as a function of the learning rate, with (a,b) exponential loss and (c,d) logistic loss on Example \ref{example:3} and \ref{example:4}. The green dashed line $\eta = \eta_0$ represents the boundary between the lazy phase and catapult phase, while black dashed line $\eta=\eta_1$ separates the catapult phase and the divergent phase. We adopt the setting of $\sigma^2_w=0.5$ and $m=100$ for exponential loss and setting for logistic loss is $\sigma^2_w=0.5$ and $m=200$. (a,c) The curves of maximum eigenvalue of NTK and Hessian coincide as predicted by the corollary \ref{cor:hessian_ntk}. (b,d) For the non-separable data, the trend of the two eigenvalue curves is consistent with the change of learning rate.}
 \label{fig:3}
\end{figure*}

We demonstrate the flatter curvature can be achieved in the {\it catapult phase} through Example \ref{example:3} and \ref{example:4}, using the code provided by \cite{nilsen2019efficient} to measure Hessian, as shown in Figure \ref{fig:3}. In the {\it lazy phase}, both curvature and eigenvalue of NTK are independent of the learning rate at the end of training. However, in the {\it catapult phase} the curvature decreases to the value smaller than that in the {\it lazy phase}. In conclusion, the NTK and Hessian have the similar behavior at the end of training on non-separable data. 

Finally, we compare our results for the catapult phase to the result with MSE loss and show the summary in Table \ref{tab:1}.

\begin{table*}[!htbp]
		\caption{A summary of the relationship between separation condition of data and the catapult phase for different losses}
		\label{tab:1}
		\begin{center}
			\begin{tabular}{|c |c | c| c|}
			\hline   
			 separation condition                                   &   linear separable                 & degenerate                  &  non-separable   \\
			\hline 
			 exponential loss (this work)                     & \xmark                               &  \cmark                & \cmark  \\
           \hline 
            logistic loss    (this work)                     &  \xmark                              & \cmark               & \cmark  \\
           \hline
		     squared loss (\cite{lewkowycz2020large})      & \cmark                               & \cmark                &\cmark       \\
           \hline 
			\end{tabular}
		\end{center}
	\end{table*}

\section{Experiment}

In this section, we present our experimental results of linear networks with the logistic loss on CIFAR-10 to examine whether fatter minima achieved in the catapult phase can lead to better generalization in the real application. We selected two ("cars" and "dogs") of the ten categories from CIFAR-10 dataset to form a binary classification problem. The results will be illustrated by comparing the generalization performance with respect to different learning rates.

\begin{figure*}[t!]
\centering
  \centering
  \includegraphics[width=1.0\textwidth]{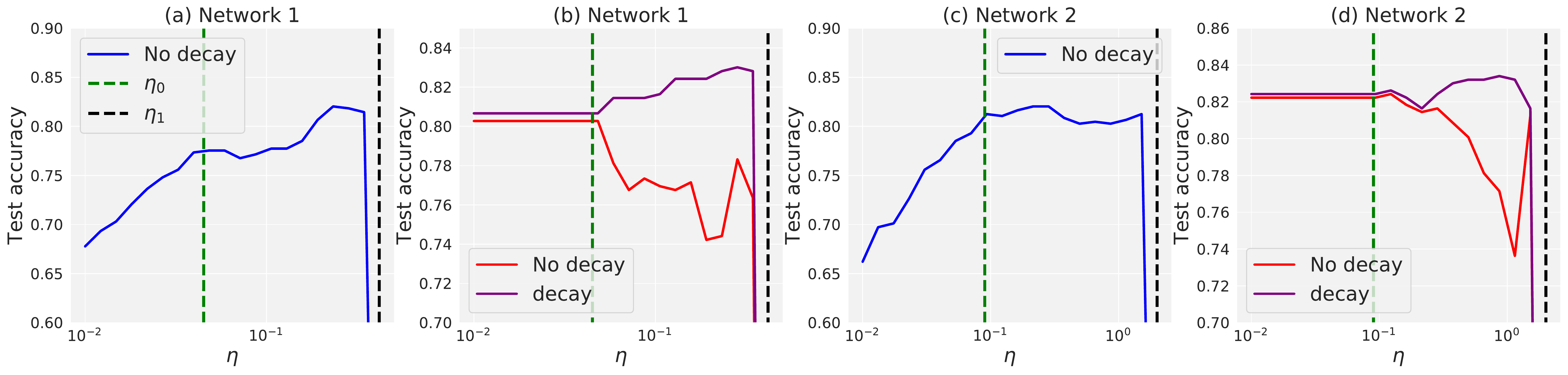}
 \caption{Test performance on CIFAR-10 dataset with respect to different learning rate phases. The data size is of $n_{\rm train}= 2048$ and $n_{\rm test} = 512$. (a,b) A two-layer linear network without bias of $\sigma_w^2 = 0.5$ and $m=500$. (c,d) A three-layer linear network with bias of $\sigma_w^2 = 0.5$, $\sigma_b^2 = 0.01$, and $m=500$. (a,c) The test accuracy is measured at the time step $t = 500$ and $t = 300$ respectively. The optimal performance is obtained when the learning rate is in the catapult phase. (b,d) The test accuracy is measured at the physical time step (red curve), after which it continues to evolve for a period of time at a small learning rate (purple): $t_{\rm phy} = 50/\eta$ and extra time $t = 500$ at $\eta = 0.01$ for the decay case. Although the results in the catapult phase do not perform as well as the lazy phase when there is no decay, the best performance can be found in the catapult phase when adopting learning rate annealing.}
 \label{fig:5}
\end{figure*}

Figure \ref{fig:5} shows the performance of two linear networks, one is one hidden layer without bias, and the other is two hidden layer linear network with bias, trained on CIFAR-10. We present results using two different stopping conditions. Firstly, we fix the training time for all learning rates, the learning rates within the catapult phase have the advantage to obtain higher test accuracy, as shown in Figure \ref{fig:5}(a,c). However, adopting a fixed training time will result in a bias in favor of large learning rates, since the large learning rate runs faster naturally. To ensure a fair comparison, we then use a fixed physical time, which is defined as $ t_{\rm phy} = t_0 \eta $, where $t_0$ is a constant. In this setting, as shown in Figure \ref{fig:5}(b,d), the performance of the large learning rate phase is even worse than that of the small learning phase. Nevertheless, we find that is achieved in the catapult phase when adopting the learning rate annealing strategy.

To explain the above experimental results, we refer to theorem 2 in \cite{ji2018gradient}. According to this theorem, the data can be uniquely partitioned into linearly separable part and non-separable part. 
When we tune the learning rate to the large learning rate regime, the algorithm quickly iterates to a flat minimum in a space spanned by non-separable data. At the same time, for linearly separable data, the gradient descent cannot achieve the maximum margin due to the large learning rate. As a result, for this part of the data, the generalization performance is suppressed. This can explain why when we fix the physical steps, the performance in the large learning rate regime is worse than that of the small learning rate phase. On the other hand, when we adopt the strategy of learning rate annealing, for non-separable data, since the large learning rate has learned a flat curvature, the subsequent small learning rate will not affect this result. For data with linearly separable parts, reducing the learning rate can restore the maximum margin. Therefore, we can see that under this strategy, the best performance can be found in the phase of large learning rate. 

\section{Discussion}

In this work, we characterize the dynamics of deep linear networks for binary classification trained with gradient descent in the large learning rate regime, inspired by the seminal work \cite{lewkowycz2020large}. We present a {\it catapult effect} in the large learning rate phase depending on separation conditions associated with logistic and exponential loss. According to our theoretical analysis, the loss in the catapult phase can converge to the global minimum like the lazy phase. However, from the perspective of Hessian, the minimum achieved in the catapult phase is fatter. We show empirically that even without SGD optimization, the best generalization performance can be achieved in the catapult stage phase for linear networks. While this work on the large learning rate for linear networks in the binary classification, there are several remaining open questions. For non-linear networks, the effect of a large learning rate is not clear in theory. In addition, the stochastic gradient descent algorithm also needs to be explored when the learning rate is large. We leave these unsolved problems for future work.

\bibliographystyle{plain}
\bibliography{ref}

\appendix

\setcounter{prop}{0}
\setcounter{thm}{0}
\setcounter{lem}{0}
\setcounter{cor}{0}

\section{Appendix} \label{sec:proofs} 
This appendix is dedicated to proving the key results of this paper, namely Proposition \ref{prop:linear_predictor} and Theorems \ref{thm:linear_predictor_degenerate}, \ref{thm:linear_network_degenerate}, and Corollary \ref{cor:hessian_ntk} which describe the dynamics of gradient descent with logistic and exponential loss in different learning rate phase. 

\begin{prop}\label{prop:linear_predictor}
For a linear predictor $f = w^T x $, along with a loss $\ell \in \{ \ell_{\exp}, \ell_{\log} \}$. 
\begin{enumerate}
\item [1] 
Under Assumption \ref{ass:1}, the empirical loss is $\beta$-smooth. Then the gradient descent with constant learning rate $\eta < \frac{2}{\beta}$ never increase the risk, and empirical loss will converge to zero:
\[
\mathcal{L}(w_{t+1}) - \mathcal{L}(w_t) \le 0, ~~~ \lim_{t \rightarrow \infty} \mathcal{L}(w_t) = 0, ~~~with~~~  \eta < \frac{2}{\beta}
\]
\item [2] 
Under Assumption \ref{ass:2}, the empirical loss is $\beta$-smooth and $\alpha$-strongly convex, where $\alpha \le \beta$. Then the gradient descent with a constant learning rate $\eta < \frac{2}{\beta}$ never increases the risk, and empirical loss will converge to a global minimum. On the other hand, the gradient descent with a constant learning rate $\eta > \frac{2}{\alpha}$ never decrease the risk, and empirical loss will explode or saturate:
\[
\begin{aligned}
&\mathcal{L}(w_{t+1}) - \mathcal{L}(w_{t}) \le 0, ~~~ \lim_{t \rightarrow \infty} \mathcal{L}(w_t) = G_0,~~~ with~~~  \eta < \frac{2}{\beta} \\
&\mathcal{L}(w_{t+1}) - \mathcal{L}(w_t) \ge 0, ~~~ \lim_{t \rightarrow \infty} \mathcal{L}(w_t) = G_1,~~~ with~~~  \eta > \frac{2}{\alpha}
\end{aligned}
\]
where $G_0$ is the value of a global minimum while $G_1 = \infty$ for exploding situation or $G_0 <G_1 < \infty$ when saturating.
\end{enumerate}
\end{prop}

\begin{proof}
\begin{enumerate}
\item [1] 
 We first prove that empirical loss $\mathcal{L}(u)$ regrading data-scaled weight $u_i \equiv w^T x_i y_i$ for the linearly separable dataset is smooth. The empirical loss can be written as $\mathcal{L}  =  \sum_{i=1}^n \ell (u_i)$, then the second derivatives of logistic and exponential loss are,
\[
    \mathcal{L}''_{\exp}  =  \sum_{i=1}^n \ell''_{\exp} (u_i)  = \sum_{i=1}^n \exp''(-u_i)  = \sum_{i=1}^n \exp(-u_i)
\]
\[
    \mathcal{L}''_{\log}  =  \sum_{i=1}^n \ell''_{\log} (u_i)  =  \sum_{i=1}^n \log''(1+\exp(-u_i)) = \sum_{i=1}^n \frac{\exp(-u_i)}{(1+\exp(-u_i))^2}
\]
when $w_t$ is limited, there will be a $\beta$ such that $\mathcal{L}''< \beta$. Besides, because there exists a separator $w_\ast$ such that $\forall i: w^T_\ast x_i y_i > 0$, the second derivative of empirical loss can be arbitrarily close to zero. This implies that the empirical loss function is not strongly convex.

Recall a property of $\beta$-smooth function $f$ \cite{bubeck2014convex},
\[
 f(y) \le f(x) + (\nabla_x f)^T(y-x) +\frac{1}{2} \beta \left\lVert y-x \right\lVert^2
\]
Taking the gradient descent into consideration,
 \[
 \begin{aligned}
 \mathcal{L}(w_{t+1}) & \le  \mathcal{L}(w_t) + \big(\nabla_{w_t} \mathcal{L}(w_t) \big)^T (w_{t+1}-w_t \big) +\frac{1}{2} \beta \left\lVert w_{t+1}-w_t \right\lVert^2 \\
                     & =  \mathcal{L}(w_t) + \big(  \nabla_{w_t}\mathcal{L}(w_t) \big)^T \big(-\eta \nabla_{w_t}\mathcal{L}(w_t)\big) +\frac{1}{2} \beta \left\lVert -\eta \nabla_{w_t}\mathcal{L} \right\lVert^2 \\ 
                     & =  \mathcal{L}(w_t) -\eta(1-\frac{\eta \beta}{2})\left\lVert  \nabla_{w_t}\mathcal{L} \right\lVert^2
\end{aligned}
 \]
when $1-\frac{\eta \beta}{2} > 0$, that is $\eta < \frac{2}{\beta}$, we have,
\[
  \mathcal{L}(w_{t+1}) \le \mathcal{L}(w_t) -\eta(1-\frac{\eta \beta}{2})\left\lVert \nabla_{w_t}\mathcal{L} \right\lVert^2 \le  \mathcal{L}(w_t)
\]
We now prove that empirical loss will converge to zero with learning rate $ \eta < \frac{2}{\beta} $. We change the form of inequality above,
\[
   \frac{ \mathcal{L}(w_t)- \mathcal{L}(w_{t+1}) }{\eta(1-\frac{\eta \beta}{2})}  \ge \left\lVert \nabla_{w_t}\mathcal{L}(w_t) \right\lVert^2
\]
this implies,
\[
 \begin{aligned}
 \sum_{t=0}^T \left\lVert \nabla_{w_t}\mathcal{L}(w_t) \right\lVert^2 & \le \sum_{t=0}^T \frac{ \mathcal{L}(w_t)- \mathcal{L}(w_{t+1}) }{\eta(1-\frac{\eta \beta}{2})} \\
                                                                        & =  \frac{ \mathcal{L}(w_0)- \mathcal{L}(w_T) }{\eta(1-\frac{\eta \beta}{2})} < \infty
 \end{aligned}
\]
therefore we have $\lim_{t \rightarrow \infty } \left\lVert \nabla_{w_t}\mathcal{L}(w_t) \right\lVert = 0$. 

\item [2]  When the data is not linear separable, there is no $w_\ast$ such that $\forall i: w^T_\ast x_i y_i > 0$. Thus, at least one $w^T_\ast x_i y_i $ is negative when the other terms are positive. This implies that the solution of the loss function is finite and the empirical loss is both $\alpha$-strongly convex and $\beta$-smooth.

Recall a property of $\alpha$-strongly convex function $f$ \cite{bubeck2014convex},
\[
 f(y) \ge f(x) + (\nabla_x f)^T(y-x) +\frac{1}{2} \alpha \left\lVert y-x \right\lVert^2
\]
Taking the gradient descent into consideration,
 \[
 \begin{aligned}
 \mathcal{L}(w_{t+1}) & \ge  \mathcal{L}(w_t) + \big(\nabla_{w_t} \mathcal{L}(w_t \big)^T \big(w_{t+1}-w_t \big) +\frac{1}{2} \alpha \left\lVert w_{t+1}-w_t \right\lVert^2 \\
                     & =  \mathcal{L}(w_t) + \big(  \nabla_{w_t}\mathcal{L}(w_t \big)^T \big(-\eta \nabla_{w_t}\mathcal{L}(w_t)\big) +\frac{1}{2} \alpha \left\lVert -\eta \nabla_{w_t}\mathcal{L} \right\lVert^2 \\ 
                     & =  \mathcal{L}(w_t) -\eta(1-\frac{\eta \alpha}{2})\left\lVert  \nabla_{w_t}\mathcal{L} \right\lVert^2
\end{aligned}
 \]
when $1-\frac{\eta \alpha}{2} < 0$, that is $\eta > \frac{2}{\alpha}$, we have,
\[
  \mathcal{L}(w_{t+1}) \ge \mathcal{L}(w_t) -\eta(1-\frac{\eta \alpha}{2})\left\lVert \nabla_{w_t}\mathcal{L} \right\lVert^2 \ge  \mathcal{L}(w_t).
\]

\end{enumerate}
\end{proof}

\begin{thm}\label{thm:linear_predictor_degenerate}
For a linear predictor $f = w^T x $ equipped with exponential (logistic) loss under assumption \ref{ass:3}, there is a critical learning rate that separates the whole learning rate space into two (three) regions. The critical learning rate satisfies $$\mathcal{L}'(w_0) = -\mathcal{L}'(w_0-\eta_{\rm critical} \mathcal{L}'(w_0)),$$ where $w_0$ is the initial weight. Moreover,  
\begin{enumerate}
\item [1] 
For exponential loss, the gradient descent with a constant learning rate $\eta < \eta_{\rm critical}$ never increases loss, and the empirical loss will converge to the global minimum. On the other hand, the gradient descent with learning rate $ \eta = \eta_{\rm critical}$ will oscillate. Finally, when the learning rate $  \eta > \eta_{\rm critical}$, the training process never decreases the loss and the empirical loss will explode to infinity:
\[
\begin{aligned}
& \mathcal{L}(w_{t+1}) - \mathcal{L}(w_t) < 0,  ~~~ \lim_{t \rightarrow \infty} \mathcal{L}(w_t) = 1,~~~ with~~~  \eta < \eta_{\rm critical}, \\
& \mathcal{L}(w_{t+1}) - \mathcal{L}(w_t) = 0,  ~~~ \lim_{t \rightarrow \infty} \mathcal{L}(w_t) = \mathcal{L}(w_0),~~~ with~~~  \eta = \eta_{\rm critical},  \\
& \mathcal{L}(w_{t+1}) - \mathcal{L}(w_t) > 0,  ~~~ \lim_{t \rightarrow \infty} \mathcal{L}(w_t) = \infty,~~~ with~~~  \eta > \eta_{\rm critical}.  \\
\end{aligned}
\]
\item [2] 
For logistic loss, the critical learning rate satisfies a condition: $\eta_{\rm critical} > 8$. The gradient descent with a constant learning rate $\eta < 8$ never increases the loss, and the loss will converge to the global minimum. On the other hand, the loss along with a learning rate $8 \le \eta <\eta_{\rm critical} $ will not converge to the global minimum but oscillate. Finally, when the learning rate $  \eta > \eta_{\rm critical}$, gradient descent never decreases the loss, and the loss will saturate:
\[
\begin{aligned}
& \mathcal{L}(w_{t+1}) - \mathcal{L}(w_t) < 0,   ~~\lim_{t \rightarrow \infty} \mathcal{L}(w_t) = \log(2),~with~ \eta < 8, \\
& \mathcal{L}(w_{t+1}) - \mathcal{L}(w_t) \le 0,   ~~\lim_{t \rightarrow \infty} \mathcal{L}(w_t) = \mathcal{L}(w_\ast) < \mathcal{L}(w_0), ~with~  8 \le \eta < \eta_{\rm critical},  \\
& \mathcal{L}(w_{t+1}) - \mathcal{L}(w_t) \ge 0,   ~~\lim_{t \rightarrow \infty} \mathcal{L}(w_t) = \mathcal{L}(w_\ast) \ge \mathcal{L}(w_0),~with~    \eta \ge \eta_{\rm critical}.  \\
\end{aligned}
\]
where $w_\ast$ satisfies $-w_\ast = w_\ast - \frac{\eta}{2} \frac{\sinh(w_\ast)}{1+\cosh(w_\ast)}$.
\end{enumerate}
\end{thm}

\begin{figure*}[t!]
	\centering
	\centering
	\includegraphics[width=0.8\textwidth]{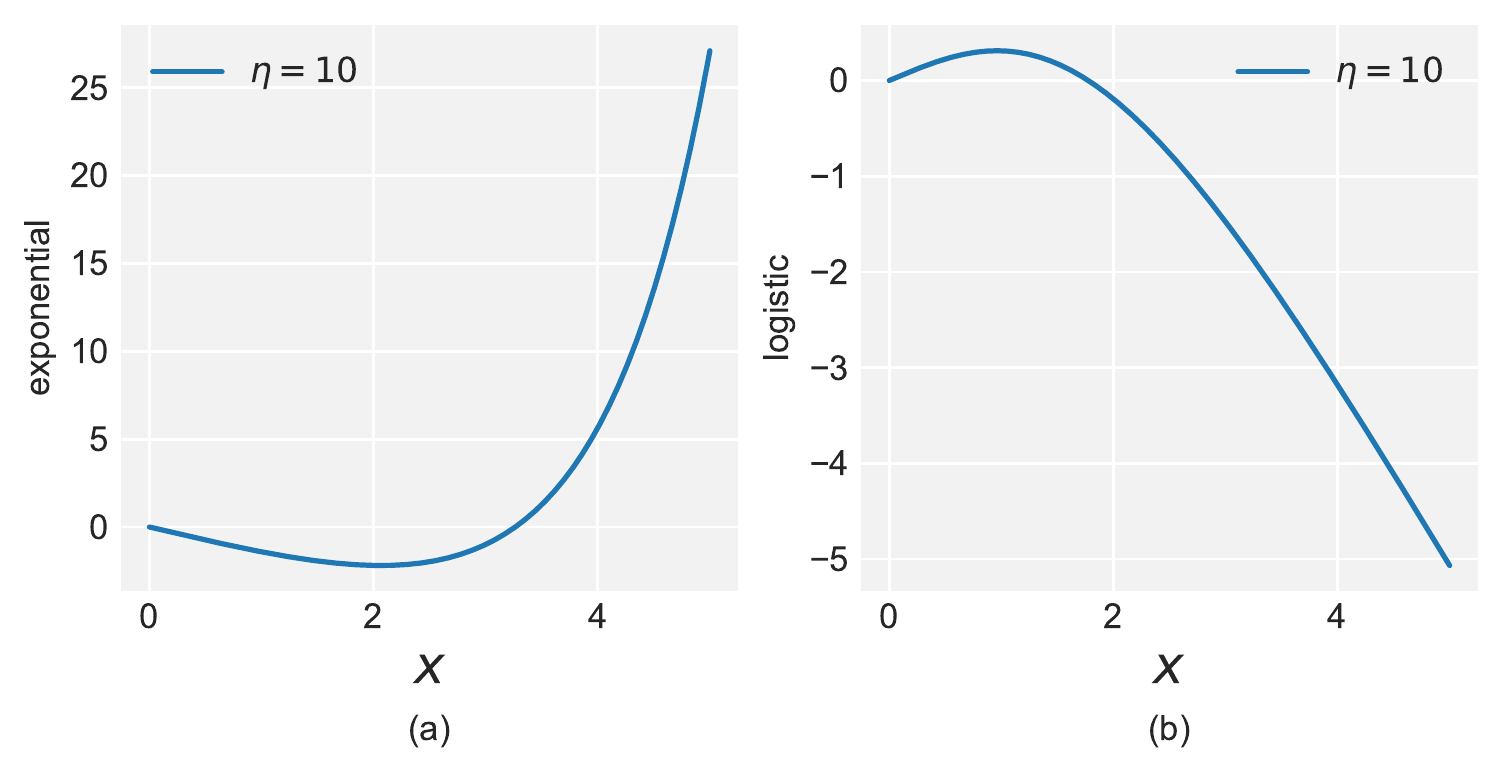}
	\caption{Graph of $\phi(x)$ for the two losses. (a) Exponential loss with learning rate $\eta = 10$. (b) Logistic loss with learning rate $\eta = 10$. }
	\label{fig:phi1}
\end{figure*}

\begin{proof}

\begin{enumerate}
\item [1] 
Under the {\it degeneracy assumption}, the risk is given by the hyperbolic function $\mathcal{L}(w_t)=\cosh(w_t)$. The update function for the single weight is,
\begin{equation} \nonumber
w_{t+1} = w_t - \eta \sinh(w_t).
\end{equation} 
To compare the norm of the gradient $\norm{\sinh (w_t)}$ and the norm of loss, we introduce the following function:
\begin{align}
\phi(x) &= \eta \mathcal{L}'(x) - 2x \\
&= \eta \sinh (x) -2x, \ \ \text{for}\ x \ge 0.
\end{align}
Then it's easy to see that
$$\mathcal{L}(w_{t+1}) > |\mathcal{L}(w_{t})| \iff \phi(|w_t|) > 0.$$
In this way, we have transformed the problem into studying the iso-surface of $\phi(x)$. Define $\ph_1$ by
$$\ph_1 = \{x | \phi(x) < 0\}.$$
Let $\ph_2$ be the complementary set of $\ph_1$ in $[0,+\infty)$. Since $\frac{\sin x}{x}$ is monotonically increasing, we know that $\ph_2$ is connected and contains $+ \infty$. 

Suppose $ \eta > \eta_{\rm critical}$, then $\phi(w_0) > 0$, which implies that
$$\mathcal{L} (w_1) > \mathcal{L} (w_0)\ \ \text{and}\ \ |w_1| > |w_0|.$$
Thus, the first step gets trapped in $\ph_2$:
$$\phi(w_1) > 0.$$
By induction, we can prove that $\phi(w_t) > 0$ for arbitrary $t \in \mathbb{N}$, which is equivalent to 
$$\mathcal{L}(w_t) > \mathcal{L}(w_{t-1}) .$$

Similarly, we can prove the theorem under another toe initial conditions: $\eta = \eta_{\rm critical}$ and $\eta < \eta_{\rm critical}$.

\item [2] 
Under the {\it degeneracy assumption}, the risk is governed by the hyperbolic function $\mathcal{L}(w_t)= \frac{1}{2}\log(2+2\cosh(w_t))$. The update function for the single weight is,
\begin{equation} \nonumber
w_{t+1} = w_t - \frac{\eta}{2} \frac{\sinh(w_t)}{1+\cosh(w_t)}.
\end{equation} 
Thus,
\begin{align}
\phi(x) &= \eta \mathcal{L}'(x) - 2x \\
&= \frac{\eta}{2}\frac{\sinh(x)}{1+\cosh(x)} - 2x, \ \ \text{for}\ x \ge 0.
\end{align}

Unlike the exponential loss, $\frac{\sinh (x)}{x(1+\cosh (x))}$ is monotonically decreasing, which means that $\ph_2$ of $\phi (x)$ doesn't contain $+ \infty$.(see figure \ref{fig:phi1}).\\
Suppose $8 < \eta < \eta_{\rm critical}$, then $w_0$ lies in $\ph_2$. In this situation, we denote the critical point that separates $\ph_1$ and $\ph_2$ by $w_\ast$. That is
$$-w_\ast = w_\ast - \eta \frac{\sinh(w_\ast)}{1+\cosh(w_\ast)}.$$
Then it's obvious that before $w_t$ arrives at $w_\ast$, it keeps decreasing and will eventually get trapped at $w_\ast$:
$$\lim_{t \rightarrow \infty} w_t = w_\ast,$$
and we have 
$$ \lim_{t \rightarrow \infty} \mathcal{L}(w_{t})-\mathcal{L}(w_{t-1})= 0.$$
When $\eta < 8$, $\ph_2$ is empty. In this case, we can prove by induction that $\phi(w_t) > 0$ for arbitrary $t \in \mathbb{N}$, which is equivalent to 
$$\mathcal{L}(w_t) > \mathcal{L}(w_{t-1}) .$$

\end{enumerate}
\end{proof}

\begin{figure*}[t!]
	\centering
	\centering
	\includegraphics[width=0.8\textwidth]{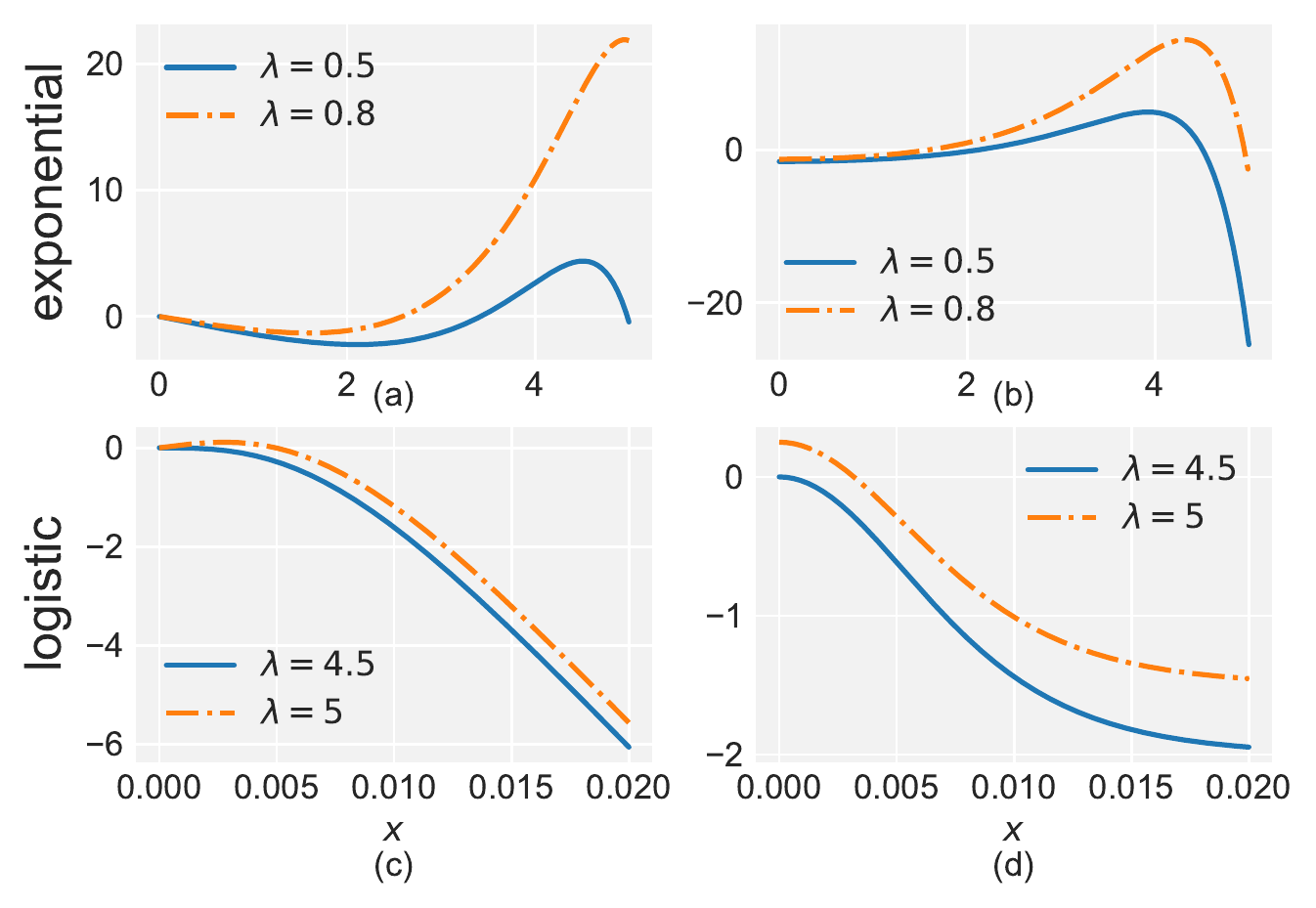}
	\caption{Different colors represent different $\lambda$(NTK) values. (a) graph of $\phi_{\lambda}(x)$ equipped with the exponential loss. (b) graph of the derivative of $\phi_{\lambda}(x)$ equipped with the exponential loss. (c) graph of $\phi_{\lambda}(x)$ equipped with the logistic loss. (d)  graph of the derivative of $\phi_{\lambda}(x)$ equipped with the logistic loss.\  Notice that the critical point of the exponential loss moves to the right as $\lambda$ decreases.}
	\label{fig:phi2}
\end{figure*}

\begin{thm}\label{thm:linear_network_degenerate}
Under appropriate initialization and assumption \ref{ass:3}, there exists a {\it catapult phase} for both the exponential and logistic loss. More precisely, when $\eta$ belongs to this phase, there exists a $T>0$ such that the output function $f_t$ and the eigenvalue of NTK $\lambda_t$ update in the following way:
\begin{enumerate}
	\item $\mathcal{L}_t$ keeps increasing when $t < T$.
	\item 
	After the $T$ step and its successors, the loss decreases, which is equivalent to:
	$$|f_{T+1}|>|f_{T+2}| \ge |f_{T+3}| \ge \dots.$$
	\item The eigenvalue of NTK keeps dropping after the $T$ steps:
	$$\lambda_{T+1} > \lambda_{T+2} \ge \lambda_{T+3} \ge \dots .$$	
\end{enumerate}
Moreover, we have the inverse relation between the learning rate and the final eigenvalue of NTK:
$	\lambda_\infty \le \lim_{t \rightarrow \infty} \frac{4f_t}{\eta \tilde {f_t}_{\rm exp}} $ with exponential loss, or $	\lambda_\infty \le \lim_{t \rightarrow \infty} \frac{4f_t}{\eta \tilde {f_t}_{\rm log} }$ with logistic loss.
\end{thm}

\begin{proof}

{\bf Exponential loss}

$\tilde f_{\rm exp}$ satisfies:
\begin{enumerate}
\item $|\tilde f_{\rm exp}(x)| = |\tilde f_{\rm exp}(-x)|$.
\item $\lim_{x \rightarrow 0} \frac{\tilde f_{\rm exp}(x)}{x} = 1.$
\item $\tilde f_{\rm exp}(x)$ has exponential growth as $x \rightarrow \infty$.
\end{enumerate}	
By the definition of the normalized NTK, we automatically get
$$\lambda_t \ge 0.$$
From the numerical experiment, we observe that at the ending phase of training, $\lambda_t$ keeps non-increasing. Thus, $\lambda_t$ must converge to a non-negative value, which satisfies
\begin{equation} \label{ending NTK}
\frac{\eta^2}{m} \lambda \tilde f^2_t - \frac{4 \eta}{m} f_t \tilde f_t \le 0.
\end{equation}
Thus, $ \lambda \le \lim_{t \rightarrow \infty} \frac{4f_t}{\eta \tilde f_t }.$\\
Since the output $f$ converges to the global minimum, a larger learning rate will lead to a lower limiting value of the NTK. As it was pointed out in \cite{lewkowycz2020large}, a flatter NTK corresponds to a smaller generalization error in the experiment. However, we still need to verify that large learning rate exists.\\
Note that during training, the loss function curve may experience more than one wave of uphill and downhill. To give a precise definition of large learning rate, it should satisfy the following two conditions:
\begin{enumerate}
\item $|f_{T+1}| > |f_T|$, this implies that
$$\mathcal{L}_{T+1} > \mathcal{L}_T.$$
For the $T+1$ step and its successors,
$$|f_{T+1}|>|f_{T+2}| \ge |f_{T+3}| \ge \dots.$$
\item The norm of NTK keeps dropping after T steps:
$$\lambda_T > \lambda_{T+1} \ge \lambda_{T+2} \ge \dots .$$	
\end{enumerate}	
If we already know that the loss keeps decreasing after the $T + 1$ step, then
\begin{equation} \label{delta lambda}
\Delta \lambda = \frac{\eta}{m} \tilde{f} \cdot (\eta \lambda \tilde f - 4f).
\end{equation}
Since $\frac{|\tilde{f}|}{|f|}\ge 1$ and is monotonically increasing when $\tilde{f} = \sinh{f}$, we automatically have

$$ \lambda_T > \lambda_{T+1} > \lambda_{T+2}\ge \dots ,$$
If $$\lambda_T < \frac{4f_T}{\eta \tilde{f}_T}\ \ and \ \ \lambda_{T+1} < \frac{4f_{T+1}}{\eta \tilde{f}_{T+1}}.$$ This condition holds if the parameters are close to zero initially.  \\To check condition (1), the following function which plays an essential role as in the non-hidden layer case:
$$\phi_{\lambda}(x) = \eta \lambda \sinh(x) - \frac{\eta^2}{m}x \sinh^2(x)  - 2x, \ \ \ \text{for}\ x \ge 0.$$
Notice that an extra parameter $\lambda$ emerges with the appearance of the hidden layer. We call it the control parameter of the function $\phi(x)$.\\
For a fixed $\lambda$, since now $\phi(x)$ becomes linear, the whole $[0,+ \infty)$ is divided into three phases (see figure \ref{fig:phi2}):
\begin{align*}
\ph_1 &:= \text{the connected component of}\   \{x|\  \phi_{\lambda}(x) < 0\} \ \text{that contains 0},\\
\ph_2 &:= \{x|\  \phi_{\lambda}(x) > 0\},\\
\ph_3 &:= \text{the connected component of}\   \{x|\  \phi_{\lambda}(x) < 0\} \ \text{that contains } + \infty.
\end{align*}
It's easy to see that $\mathcal{L}_{\rm exp} (f_{T+1}) > \mathcal{L}_{\rm exp} (f_T)$ if and only if
$$\phi_{\lambda_T}(f_T) > 0.$$
That is, $f_T$ lies in $\ph_2$ of $\phi_{\lambda_T}$. Similarly,
$$\mathcal{L}_{\rm exp}(f_{T+2}) < \mathcal{L}_{\rm exp}(f_{T+1}) \ \iff\ \phi_{\lambda_{T+1}}(f_{T+1}) < 0.$$
That is, $f_{T+1}$ jumps into $\ph_1$ of $\phi_{\lambda_{T+1}}$. Denote the point that separates $\ph_1$ and $\ph_2$ by $x_\ast$, then form the graph of $\phi_{\lambda}(x)$ with different $\lambda$, we know that
$$x_\ast (\lambda') > x_\ast (\lambda)\  \ \text{if}\ \ \lambda' < \lambda.$$
Therefore, condition (1) is satisfied if
\begin{equation}
\label{3}x_\ast (\lambda_{T+1}) > f_T + \phi_{\lambda_T}(f_T)
\end{equation}
and at the same time,
$$\lambda_{T+1} - \lambda_T > 0.$$
For simplicity, we reset $T$ as our initial step. Write the output function $f_t$ as
\begin{equation}
\begin{aligned}
f_{t+1} = f_t(1+\mathcal{A}_t),
\end{aligned}
\end{equation}
where $\mathcal{A}_t =  \frac{\eta^2}{m}\tilde f^2_t - \eta \lambda_t \tilde f_t /f_t$. Thus, $\phi_{\lambda_0}(f_0) > 0$ is equivalent to $\mathcal{A}_0 < -2$.\\
Similarly, write the update function for $\lambda_t$ as
\begin{equation}
\begin{aligned}
\lambda_{t+1} = \lambda_t(1+\mathcal{B}_t),
\end{aligned}
\end{equation}
where $\mathcal{B}_t = \frac{\eta^2}{m} \tilde f^2_t - \frac{4\eta}{m} \tilde f_t f_t/\lambda_t$. To fulfill the above condition on NTK, we need
$$\mathcal{B}_0 < 0.$$
To check (\ref{3}), let the initial output $f_0$ be close to $X_\ast(\lambda_{0})$ (this can be done by adjusting $w_0$):
$$0 < f_0 - X_\ast < \epsilon.$$
Then by the mean value theorem,
$$x_\ast (\lambda_1) - x_\ast (\lambda_0) = \frac{\partial x_\ast }{\partial \lambda_{\ast}}(\lambda_{\ast}) \cdot \Delta \lambda.$$
The derivative $\frac{\partial x_\ast }{\partial \lambda_{\ast}}$ can be calculated by the implicit function theorem:
\begin{align*}
\frac{\partial x_\ast }{\partial \lambda} &= - \frac{\partial \phi_{\lambda}(x_\ast)}{\partial \lambda}/ \frac{\partial \phi_{\lambda}(x_\ast)}{\partial x_\ast} \\
& = - \eta \sinh (x_\ast) / \frac{\partial \phi_{\lambda}(x_\ast)}{\partial x_\ast} .
\end{align*}
It's easy to see that $|\frac{\partial x_\ast }{\partial \lambda}|$ is bounded away from zero if the initial output is in $\ph_2$ and near $x_\ast$ of $\phi_{\lambda_0}(x)$ (see Figure \ref{fig:phi2}).\\
On the other hand, we have the freedom to move $f_0$ towards $x_\ast$ of $\phi_{\lambda_0}(x)$ without breaking the $\Delta \lambda < 0$ condition. Since
$$|\frac{\eta \lambda \tilde{f}}{4f}| < |\frac{\eta \lambda \tilde{f}'}{4f'}|\ \ \ \text{if}\ \ f < f'.$$
Therefore, we can always find a $\epsilon > 0$ such that $0 < f_0 - x_{\ast} < \epsilon$ and (\ref{3}) is satisfied. Combining the above, we have demonstrated the existence of the {\it catapult phase}  for the exponential loss. \\

{\bf logistic loss}

When Considering the degeneracy case for the logistic loss, the loss will be 
\begin{equation}
\mathcal{L} =  \frac{1}{2}\log(2+2\cosh(m^{-1/2} w^{(2)} w^{(1)})).
\end{equation}
Much of the argument is similar. For example, Equation (\ref{ending NTK}) still holds if we replace $\tilde{f}_{\rm exp}$ by
$$\tilde{f}_{\rm log}(x) : = \frac{\sinh (x)}{1 + \cosh(x)}.$$
$\tilde{f}_{\rm log}$ satisfies
\begin{enumerate}
\item $|\tilde f_{\rm log}(x)| = |\tilde f_{\rm log}(-x)|$.
\item $|\tilde f_{\rm log}(x)| \leq 1\ \ \ \text{for} ~x \in (-\infty,\infty).$ 
\end{enumerate}	
This implies that $$|\frac{\tilde{f}}{f}| \leq \frac{1}{2}.$$ Then by (\ref*{delta lambda}), we have
$\Delta \lambda < 0$ if $\lambda \leq \frac{8}{\eta}$. Thus, condition 2 is satisfied for both loss functions. Now, $\phi_{\lambda}(x)$ becomes:
$$\phi_{\lambda}(x) : = \eta \lambda \frac{\sinh (x)}{1 + \cosh (x)} -\frac{\eta^2}{m}x \frac{\sinh^2(x)}{(1+\cosh (x))^2} - 2x,$$
along with its derivative:
\begin{align*}
\phi_{\lambda}'(x) : = &\eta \lambda \frac{\cosh (x)}{1 + \cosh (x)} - \frac{\eta \lambda \sinh^2(x) }{(1+\cosh (x))^2} - 2 \\
-&  \frac{\eta^2}{m} \frac{\sinh^2(x) }{(1+\cosh (x))^2} - 2 \frac{\eta^2}{m}\frac{\sinh (x)}{1 + \cosh (x)} \big[\frac{\cosh (x)}{1 + \cosh(x)} - \frac{\sinh^2(x)}{(1+\cosh (x))^2} \big].  
\end{align*} The method of verifying condition 1 is similar with the exponential case, except that $\phi_{\lambda}(x)$ has only $\ph_1$ and $\ph_2$ (see figure \ref{fig:phi2}). As the NTK $\lambda$ goes down, $\ph_1$ will disappear and at that moment the loss will keep decreasing. Let $\lambda_*$ be the value such that $\phi_{\lambda_*}'(x) = 0$, then
$$\lambda_* = 4/\eta .$$
During the period when $4/\eta < \lambda_{t} < 8/\eta$, the NTK keeps dropping and the loss may oscillate around $x_{*}$. However, we may encounter the scenario that both the loss and the $\lambda_{t}$ are going up before dropping down simultaneously (see the first three steps in figure \ref{fig:phi2}). Theoretically, it corresponds to jump from $\ph_2$ to $\ph_3$ and then to $\ph_1$ of $\phi_{\lambda_{1}}(x)$ in the first two steps.
This is possible since $\tilde{f}_{\rm log}$ is decreasing when $x > 0$. This implies that
$$|\frac{\eta \lambda \tilde{f}'}{4f'}| < |\frac{\eta \lambda \tilde{f}}{4f}|\ \ \ \text{if}\ \ f < f'.$$
So an increase of the output will cause the NTK to drop faster.
\end{proof}

\begin{cor} [Informal] \label{cor:hessian_ntk}
Consider optimizing $\mathcal{L}(w)$ with one hidden layer linear network under assumption \ref{ass:3} and exponential (logistic) loss using gradient descent with a constant learning rate. For any learning rate that loss can converge to the global minimal, the larger the learning rate, the flatter curvature the gradient descent will achieve at the end of training.
\end{cor}

\begin{proof}
The Hessian matrix is defined as the second derivatives of the loss with respect to the parameters,
$$ H_{\alpha \beta} = \frac{\partial^2 \mathcal{L}}{\partial \theta_\alpha  \partial \theta_\beta}$$
where $\theta_\alpha,\theta_\beta \in \{w^{(1)}, w^{(2)} \}$ for our setting of linear networks. For logistic loss,
\begin{align*}
 H_{\alpha \beta} & = \frac{1}{n} \sum_i  \frac{\partial^2 \exp(-y_i f_i)}{\partial \theta_\alpha  \partial \theta_\beta} \\
                  & = \frac{1}{n} \sum_i  \big[ \frac{\partial^2 f_i}{\partial \theta_\alpha  \partial \theta_\beta} \exp(-y_i f_i)(-y_i)
                                                +\frac{\partial f_i}{\partial \theta_\alpha}\frac{\partial f_i}{\partial \theta_\beta} \exp(-y_i f_i) \big]\\
\end{align*}
We want to make a connection from Hessian matrix to the neural tangent kernel. Note that the second term contains $\frac{\partial f_i}{\partial \theta_\alpha}\frac{\partial f_i}{\partial \theta_\beta}$, which can be written as $J J^T$, where $J = {\rm vec}[\frac{\partial f_i}{\partial \theta_j}]$. While NTK can be expressed as $J^T J$. It is known that they have the same eigenvalue. Further more, under assumption \ref{ass:3}, we have $n=2$ and $f_1 = f_2 $, thus,
\begin{align*}
 H_{\alpha \beta} & = \frac{1}{n} \sum_i \big[ \frac{\partial^2 f_i}{\partial \theta_\alpha  \partial \theta_\beta} \frac{\partial \mathcal{L}}{\partial f_\theta}
                                       +\frac{\partial f_i}{\partial \theta_\alpha}\frac{\partial f_i}{\partial \theta_\beta} \mathcal{L} \big]\\
\end{align*}
Suppose at the end of gradient descent training, we can achieve a global minimum. Then we have, $\frac{\partial \mathcal{L}}{\partial f_\theta} = 0$, and $\mathcal{L} = 1$. Thus, the Hessian matrix reduce to,
$$ H_{\alpha \beta}  = \frac{1}{n} \sum_i \frac{\partial f_i}{\partial \theta_\alpha}\frac{\partial f_i}{\partial \theta_\beta}$$
In this case, the eigenvalues of Hessian matrix are equal to those of neural tangent kernel. Combine with the \ref{thm:linear_network_degenerate}, we can prove the result.  

For logistic loss, 
\begin{align*}
 H_{\alpha \beta} & = \frac{1}{n} \sum_i  \frac{\partial^2 \log(1+\exp(-y_i f_i))}{\partial \theta_\alpha  \partial \theta_\beta} \\
                  & = \frac{1}{n} \sum_i  \big[ \frac{\partial^2 f_i}{\partial \theta_\alpha  \partial \theta_\beta} \frac{\exp(-y_i f_i)(-y_i)}{1+\exp(-y_i f_i)}
                                                +\frac{\partial f_i}{\partial \theta_\alpha}\frac{\partial f_i}{\partial \theta_\beta} \frac{\exp(-y_i f_i)}{(1+\exp(-y_i f_i))^2} \big]\\
\end{align*}
 Under assumption \ref{ass:3}, we have $n=2$ and $f_1 = f_2 $, thus,
\begin{align*}
 H_{\alpha \beta} & = \frac{1}{n} \sum_i \big[ \frac{\partial^2 f_i}{\partial \theta_\alpha  \partial \theta_\beta} \frac{\partial \mathcal{L}}{\partial f_\theta}
                                       +\frac{\partial f_i}{\partial \theta_\alpha}\frac{\partial f_i}{\partial \theta_\beta} \frac{\exp(-y_i f_i)}{(1+\exp(-y_i f_i))^2}  \big]\\
\end{align*}
Suppose at the end of gradient descent training, we can achieve a global minimum. Then we have, $\frac{\partial \mathcal{L}}{\partial f_\theta} = 0$, and $f_i = 0$. Thus, the Hessian matrix reduce to,
$$ H_{\alpha \beta}  = \frac{1}{4n} \sum_i \frac{\partial f_i}{\partial \theta_\alpha}\frac{\partial f_i}{\partial \theta_\beta}$$
In this case, the eigenvalues of Hessian matrix and NTK have the relation $\frac{1}{4} \lambda_{\rm NTK} = \lambda_{\rm Hessian} $.

\end{proof}

\end{document}